\documentclass[sigconf, timestamp=false]{acmart}

\usepackage{booktabs,tcolorbox,epsfig,balance} 
\definecolor{ao}{rgb}{0.0, 0.5, 0.0}
\usepackage{color}

\copyrightyear{2018}
\acmYear{2018} 
\setcopyright{iw3c2w3}
\acmConference[WWW 2018]{The 2018 Web Conference}{April 23--27, 2018}{Lyon, France}
\acmBooktitle{WWW 2018: The 2018 Web Conference, April 23--27, 2018, Lyon, France}
\acmPrice{}
\acmDOI{10.1145/3178876.3186053}
\acmISBN{978-1-4503-5639-8/18/04}

\begin{document}
\title{Multi-Task Pharmacovigilance Mining from Social Media Posts}

\author{Shaika Chowdhury}
\affiliation{%
  \institution{Department of Computer Science}
  {University of Illinois at Chicago}
  \city{Chicago} 
  \state{IL} 
  \postcode{60607}
}
\email{schowd21@uic.edu}

\author{Chenwei Zhang}
\affiliation{%
  \institution{Department of Computer Science}{University of Illinois at Chicago}
  \city{Chicago} 
  \state{IL} 
  \postcode{60607}
}
\email{czhang99@uic.edu}

\author{Philip S. Yu}
\affiliation{%
  \institution{Department of Computer Science}{University of Illinois at Chicago}
  \city{Chicago} 
  \state{IL} 
  \postcode{60607}
}
\email{psyu@uic.edu}

\renewcommand{\shortauthors}{S. Chowdhury et al.}

\begin{abstract}
Social media has grown to be a crucial information source for pharmacovigilance studies where an increasing number of people post adverse reactions to medical drugs that are previously unreported. Aiming to effectively monitor various aspects of Adverse Drug Reactions (ADRs) from diversely expressed social medical posts, we propose a multi-task neural network framework that learns several tasks associated with ADR monitoring with different levels of supervisions collectively.  Besides being able to correctly classify ADR posts and accurately extract ADR mentions from online posts, the proposed framework is also able to further understand reasons for which the drug is being taken, known as `indications', from the given social media post. A \textit{coverage-based} attention mechanism is adopted in our framework to help the model properly identify `phrasal' ADRs and Indications that are attentive to multiple words in a post.  Our framework is applicable in situations where limited parallel data for different pharmacovigilance tasks are available. We evaluate the proposed framework on real-world Twitter datasets, where the proposed model outperforms the state-of-the-art alternatives of each individual task consistently.
\end{abstract}

%
%



\maketitle

\section{Introduction}
Various prescription drugs intended to be taken for medical treatment are released into the market. However, studies have found that many of such drugs can be counterproductive \cite{chee2011predicting, patki2014mining}. The harmful reactions or injuries that are caused by the intake of drugs are known as Adverse Drug Reactions (ADR) \cite{sarker2015utilizing} and happens to be the fourth leading cause of death in the United States \cite{chee2011predicting}. Preventive steps such as ADR monitoring and detection, which are collectively called Pharmacovigilance, are critical to ensure safety to patients' health. Initial activities toward pharmacovigilance during clinical trials fail to reliably signal all the negative effects that could potentially be caused by a drug due to certain restrictions, calling the need for post-market ADR surveillance. Nevertheless, potentially harmful drugs remain unflagged as traditional post-market ADR monitoring methods suffer from under-reporting, incomplete data, and delays in reporting \cite{sarker2015utilizing}.

Owing to the presence of large user base on social media, a vast amount of data gets generated. Compared to clinical information retrieval from electronic health records (EHR) that pose the problem of limited access \cite{wu2017intrainstitutional}, the free accessibility of web data presents a lucrative source of medical data. This data has recently caught the attention of researchers for public health monitoring\cite{leaman2010towards, paul2011you}. Task-oriented crowdsourcing platforms are designed to collect patient feedbacks as introduced in \cite{peleg2017crowdsourcing}. A popular platform where health-related data in the form of posts/ tweets are exchanged between users is Twitter\footnote{https://twitter.com}. The posts cover a wide range of topics concerning health such as experience getting a disease/illness, symptoms of it, the drugs that were taken and harmful reactions from them.

Social media platforms offer a robust source of health-related data that patients are found to consult to learn about other's shared health-related experiences \cite{sarker2015utilizing,lampos2015assessing,zou2016infectious,yom2017predicting}. The fact that this data is up-to-date and is generated by patients overcomes the weaknesses of traditional ADR surveillance techniques. Thus social media is indispensably important and could complement traditional information sources for more effective pharmacovigilance studies, as well as potentially serve as an early warning system for unknown ADRs \cite{nikfarjam2011pattern}. However, users with different background knowledge or various linguistic preferences tend to generate social media posts with diverse expressions and ambiguous mentions, which pose a series of challenges for pharmacovigilance studies:

\begin{itemize}
\item{\textbf{Diversely expressed ADRs}}: In Twitter, the use of non-medical terms to craft one's tweet is very prevalent. For example, \textit{never sleeping} or \textit{sleep deprived} being used to describe the ADR \textit{insomnia}. These `phrasal' ADRs framed with casual words could easily merge with other irrelevant words in the post and make the detection task more difficult. 
\item{\textbf{Indications misidentified as ADRs}}: An `Indication' in a health-related post can be defined as a medical condition, disease or illness for which the drug has been prescribed to treat it. As both indications and ADRs can be referred to by the same drug, indications could be easily misidentified for ADRs. Without a deep understanding of the role that each symptom plays in a post, such misidentification could be problematic.
\end{itemize}

\begin{figure}[h]
\begin{tcolorbox}
\textit{Tweet 1}:
I'm starting to think my \textcolor{red}{Paroxetine} turns \textcolor{cyan}{panic attacks} into \textcolor{ao}{fat}.\newline
\textit{Tweet 2}:
Any ideas on treating \textcolor{cyan}{depression} naturally \textcolor{red}{Paxil} is making me \textcolor{ao}{gain unwanted pounds} and \textcolor{ao}{water weight}...Thanks
\end{tcolorbox}
\caption{Pharmacovigilance reported in social media posts.}\label{fig::sample_tweets}
\end{figure}

Figure \ref{fig::sample_tweets} shows common tweets for Pharmacovigilance taken from PSB 2016 Social Media Shared Task Twitter dataset \cite{sarker2016social}, where the token in red denotes the `Drug' being taken, blue denotes the `Indication', and green the `ADR'.
In Tweet 1, the drug \textit{Paroxetine} is being taken to treat \textit{panic attacks}, with an adverse reaction of weight gain mentioned as \textit{fat}. So \textit{panic attacks} is an instance of Indication while \textit{fat} is the ADR. In Tweet 2 while the drug \textit{Paxil} is prescribed to treat the indication \textit{depression}, it causes ADRs of \textit{gain unwanted pounds} and \textit{water weight}. In both cases it is possible for the Indications 'panic attacks' and 'depression' to get mislabeled as both commonly occur as ADRs as well. 

All previous works \cite{cocos2017deep,huynh2016adverse,nikfarjam2011pattern,nikfarjam2015pharmacovigilance} in pharmacovigilance studies have focused on solving the ADR tasks separately. As a tweet containing ADR mention can also have other medical mentions such as Indications or beneficial effects, it is important to incorporate these representations to ensure disambiguity when trying to learn each task. Lexicon-based approaches \cite{leaman2010towards, benton2011identifying, yang2012social} have been commonly adopted in earlier approaches, where ADR tokens in a text fragment are looked up in a corpus containing ADR mentions. However, the occurrence of non-medical terminologies to describe ADRs in social media makes them unsuitable. While machine learning approaches such as Naive Bayes, Support Vector Machine and Maximum Entropy \cite{bian2012towards, yang2013identification, jiang2013mining} require the use of hand-engineered features.

Trying to conquer these problems, we propose a multi-task framework based on the sequence learning model that jointly learns several related pharmacovigilance tasks. Mining from pharmacovigilance forms an interesting multi-task problem as complimentary pharmacovigilance tasks with supervision at different levels can be jointly learned. The data for these related tasks share semantic and syntactic similarities, and leveraging these shared representations can improve learning efficiency and prediction accuracy. The joint learning of the objective functions of the tasks can transfer what is learned from one task to other tasks and improve them. This is suitable for pharmacovigilance tasks as they contain data with occurrences of multiple medical terms, which otherwise makes identification of a certain category (i.e ADR or Indication) of medical term difficult. We incorporate these ideas by extending the basic recurrent neural network encoder-decoder such
that our multi-task model shares an encoder among all the tasks and
uses a different decoder for each task. 
Our assumption is that the shared encoder would learn predictive representations capturing the nuances of each task and, hence, help disambiguate an 'ADR' from an 'Indication'. On the other hand, a task-specific decoder decodes the shared encoder representation to produce task-specific output. Furthermore, by incorporating multi-granular supervision through different tasks, the decoder can successfully produce output at the sentence and word level.
\begin{figure*}
\epsfig{file=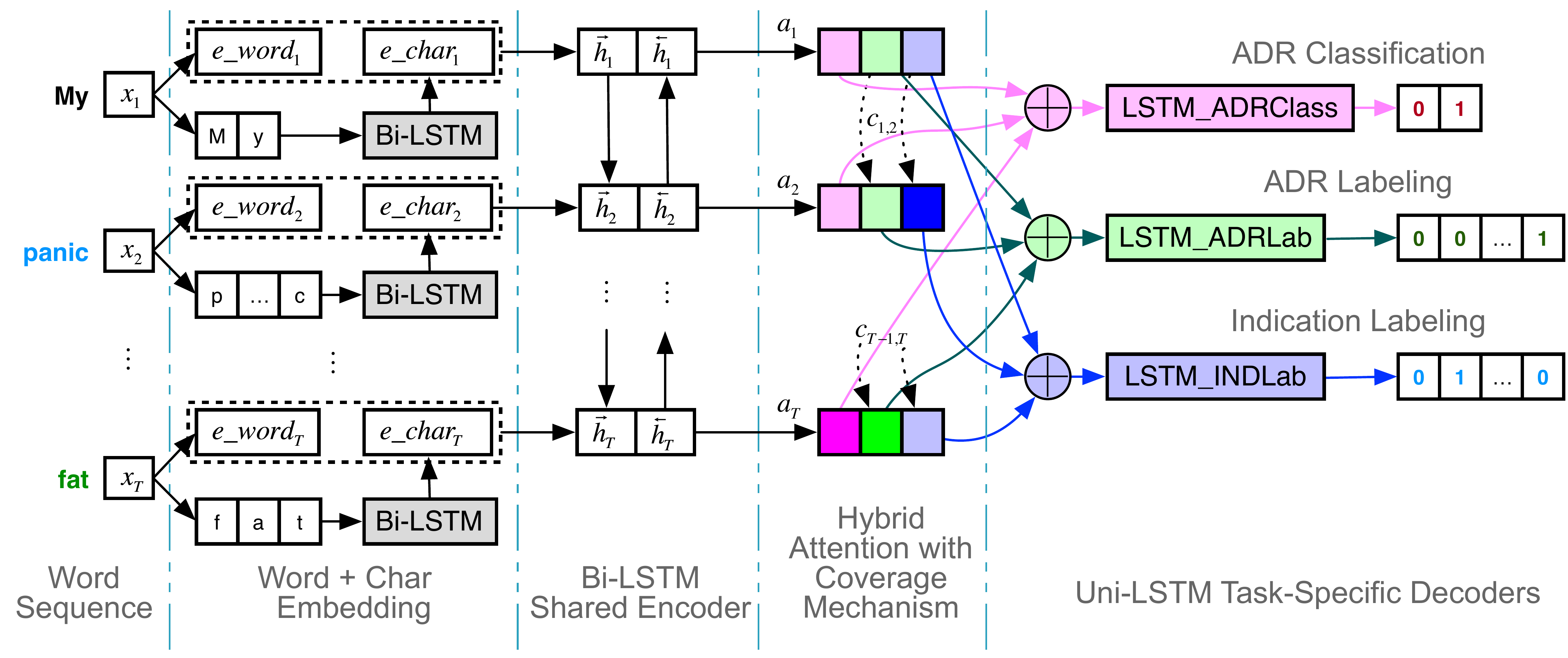, width=5.6in}
\caption{Model Architecture. We use different colors for each task-specific decoder. The shade of the colored block indicates its value. For example, the second word \textit{panic} is an Indication, so gets higher attention weight (the dark blue block), which helps in successfully labeling it as an Indication.}\label{fig::architecture}
\end{figure*}

Recent years have seen a spike in the use of sequence-to-sequence models for difficult learning tasks like machine translation \cite{sutskever2014sequence} and summarization \cite{nallapati2017summarunner}. A sequence-to-sequence model takes a sequence as input into an encoder, projects it to an intermediate encoded representation, which is then passed through a decoder to generate a sequence as the output. The proposed architecture shown in Figure \ref{fig::architecture} is a recurrent neural network-based encoder-decoder model augmented with attention and coverage mechanism to handle multiple tasks for pharmacovigilance. The reason for using a sequence to sequence model to model the multiple tasks is to be able to capture their shared representations with the encoder and generate outputs for each task with multiple decoders. We propose a multi-task learning framework for three pharmacovigilance tasks --- ADR Classification, ADR Labeling and Indication Labeling--- hypothesizing that the interactions between the tasks through joint learning would improve learning and generalization ability of each individual task. While most works on ADR classification and detection have tried to learn a single objective function, by jointly learning multiple objective functions for the tasks we introduce a novel approach for pharmacovigilance that is seen to boost the performance. Moreover, the learned features from multiple tasks help to reduce the false positives in mislabeling Indications as ADRs and vice versa. When an ADR\textbackslash{}Indication occurs as a phrase rather than a single word, it can make the detection task more difficult. Adding \textit{coverage} to the attention mechanism helps overcome this as it accumulates the attention from all the previous decoder time steps and facilitates in learning all previous ADR\textbackslash{}Indication words in a phrase. 

The main contributions of this work are summarized as follows:
\begin{enumerate}
\item Designed a unified machine learning framework to learn several pharmacovigilance tasks from social media posts simultaneously. To the best of our knowledge, this problem has not been studied carefully and has scope for novel research.
\item Adding \textit{coverage} to attention mechanism has shown to improve detection of not only 'phrasal', but also single worded ADRs and Indications.
\item State-of-the-art in terms of results obtained on real-world Twitter datasets compared to contemporary work in pharmacovigilance.
\end{enumerate}
\section{Preliminary: Tasks}
With our multi-task framework, we jointly learn three pharmacovigilance tasks, where the input tweet representation is encoded through a shared encoder. Each task is modeled as a sequence classification or sequence labeling problem. Description of each task is given below.
\subsection{ADR Classification}
This is a binary classification task to separate out the ADR assertive posts. The two classes are `ADR' and `NotADR' where `ADR' label indicates a post with ADR mention in it. While `NotADR' means it does not have any ADR, although can have other medical terms such as drugs and Indications. 

It tries to learn a function that maps a sequence $\textbf{x}$ to a class label $l\in R^L$ where $L$ is the total number of classes \cite{xing2010brief} and $L=2$. 
\begin{equation}
\textbf{x} \to l, l\in R^L
\end{equation}

\subsection{ADR Labeling}
A sequence labeling task aiming to identify ADR in a post. The detected ADRs are tagged with the `ADR' label. It tries to find the most likely sequence of tags given the input sequence $\textbf{x}$, that is the sequence with the highest probability.
\begin{equation}
{\mathbf{y}}' = \mathop {\operatorname{argmax} }\limits_y P(y|x)
\end{equation}

We use two tags to annotate the tokens of the input for ADR labeling. 'ADR' tag corresponds to a token with ADR mention and 'O' tag denotes a non-ADR word.

\subsection{Indication Labeling}
A sequence labeling task aiming to identify Indication in a post. The detected Indications are tagged with the `Indication' label. Like ADR labeling, it tries to find the tagged sequence of highest probability.

We used two tags to annotate the tokens of the input for Indication labeling. `IND' tag corresponds to a token with Indication mention and `O' tag denotes a non-Indication word.

\section{Our Model}
Our proposed multi-task framework is depicted in Figure \ref{fig::architecture}. It is composed mainly of three components --- embedding module, encoder and decoder.  The embedding module is intended to capture the meaning and semantic associations between pharmacovigilance words. All the tasks have a common encoder so that shared representations can be generated, capturing both the ADR and Indication contextual information. Lastly, the decoders employ a combined attention and coverage mechanism to facilitate the detection of ADR and Indications of various lengths. Each component of our RNN Encoder-Decoder model is described in detail in the following subsections. 
\subsection{Input}
Our input $\textbf{x} = \left(x_1, x_2,..., x_T\right)$ is a piece of text corresponding to a social media post comprising a sequence of $T$ words, where each $x_i$ represents a word in the vocabulary of size $V$.

\subsection{Word Representations}
Medical terms in a tweet post can take different roles and the same medical word or phrase can mean different things in different context. For example, `panic attacks' can occur both as an ADR as well as an Indication depending on the mention of the drug and other patterns. In order to capture their meanings as well as the semantic relationships and context they are being used in, we generate their word embeddings.

The character representation of each word is also generated to capture its morphological features. These character representations can help in capturing the representations for words such as `sleeping' where the word embedding matrix might have an entry for just `sleep'. This character representation is similar to that implemented for doing Named Entity Recognition \cite{rei2016attending}.  

The final representation of a word is the concatenation of its word embedding and character representations to incorporate users' diverse expressions.

\textbf{Word Embedding}: For each word $x_t$ we get its corresponding low-dimensional dense vector, $e\_{word_t}$, by looking up in a V x m size word-embedding matrix, where m is the dimensions of the word-embedding. 

\textbf{Character Representation}: Each word in a sentence can be denoted as  $x_t = \left(c1, c2,...\right)$ such that $c_t\in R^G$, where G is the vocabulary size of all the characters. Similar to word embedding, we first get the character embedding vector for each character by looking up in a G x p character-embedding matrix, where p is the dimensions of the character-embedding. The sequence of character-embedding vectors, $e\_{char_t}$, of the word is then fed to a bidirectional LSTM \cite{hochreiter1997long}. The final character representation, $e'\_{char_t}$, is obtained by concatenating the forward and backward final states.  

The final word embedding vector $e_t$ for each word is thus, 
\begin{equation}
e_t = \left[e\_{word_t}, e'\_{char_t}\right]
\end{equation}

\subsection{Encoder}
We use a single layer Bi-Directional RNN as the encoder, with LSTM as the basic recurrent unit due to its ability to incorporate long-term dependencies \cite{hochreiter1997long}. As the purpose of the encoder is to capture the shared representations of the multiple tasks, these representations should include contextual information in both the forward and backward directions, where the outputs of each task can depend on these previous and future elements in the sequence. So we pass the input sequence through a Bi-LSTM to serve this purpose. The Bi-LSTM achieves this by passing the input sequence in its original order through a forward LSTM, which encodes a hidden state $\overrightarrow{h}_t$ for each time step t. Also, a reversed copy of the input sequence is passed through a backward LSTM, that encodes a hidden state $\overleftarrow{h}_t$. The forward and backward hidden states are then concatenated to represent the final hidden state of the encoder at each time step.

\begin{equation}
h_t = \left[\overrightarrow{h}_t, \overleftarrow{h}_t\right]
\end{equation}

\begin{equation}
\overrightarrow{h}_t = LSTM\left[x_t, \overrightarrow{h}_{t-1}\right]
\end{equation}

\begin{equation}
\overleftarrow{h}_t = LSTM\left[x_t, \overleftarrow{h}_{t-1}\right]
\end{equation}

\subsection{Decoder}
On the decoder side, we allocate a single layer uni-directional LSTM with attention mechanism for each task in order to produce output specifically for that task. The \textit{Coverage} mechanism is integrated in the attention mechanism to give the model a sense of how much attention it has already assigned so far.

\subsubsection{Attention Mechanism}
Conventional Encoder-Decoder for sequence-to-sequence tasks has shown to have limitations as it tries to decode a fixed-length encoded vector when generating output \cite{bahdanau2014neural}. The fact that the last encoder hidden state is required to hold the summary of all the timesteps, in practice, doesn't hold, especially, for longer sentences \cite{cho2014properties}.

The use of attention has come a long way in tasks ranging from image captioning \cite{vinyals2015show} to machine translation \cite{cho2014properties}. By peeking over all the encoder states and weighing them according to their relevance to the current output to generate, it produces the \textit{attention distribution}. This attention distribution gives a signal where to attend to more in the input sequence.

We use additive attention from \cite{bahdanau2014neural}, where \textit{attention distribution} $a_t$ is calculated as:

\begin{equation}
{\textit{g}_t}_i = {\textit{v}_a^T}tanh\left(W_ah_i + W_bs_{t-1} + b_{attn}\right)
\end{equation}

\begin{equation}
a_t = softmax\left({\textit{g}}_t\right)
\end{equation}

$h_i$ is an encoder hidden state and $s_{t-1}$ is the previous decoder hidden state at decoding timestep t. $v_a$, $W_a$, $W_b$, and $b_{attn}$ are learnable attention parameters. Finding an attention score using the attention function ${\textit{g}_t}_i\left(h_i, s_{t-1}\right)$ between each encoder state $h_i$ and the decoder state preceding the current output state, tells exactly where to pay attention to in the input sequence when decoding.

The encoder hidden states are then combined with the attention distribution to give the \textit{context vector} defined as the attention weighted sum of the encoder hidden states,

\begin{equation}
con_t = \sum\nolimits_{i} h_i{a_t}_i
\end{equation}

This context vector, along with the previous decoder hidden state $s_{t-1}$, outputs the label generated at previous timestep, $y_{t-1}$. The aligned encoder hidden state $h_t$ is used to compute the decoder state $s_t$ at time step t for ADR and Indication labeling tasks.

\begin{equation}
s_t = \textit{f}\left(s_{t-1},h_t, y_{t-1}, {con}_t\right)
\end{equation}

Whereas for the ADR Classification task, the decoder state at time step t is a function of $s_{t-1}$ and ${con}_t$. 

\begin{equation}
s_t = \textit{f}\left(s_{t-1}, {con}_t\right)
\end{equation}

For all the tasks, initial decoder hidden state $s_0$ is set to the final encoder hidden state. 

\subsubsection{Coverage Mechanism}

Due to the colloquial nature of conversations on social network, ADRs are expressed in everyday language and also take phrasal form. Consider the following example tweet, 
\textit{Just took Seroquel. Now I'm freaking out that I will end up sleeping 15 hrs and miss my 12pm appt tomorrow},
where \textit{sleeping 15 hrs} is an ADR taking a phrase form. The phrasal ADRs can also be expressed as a list of ADRs, illustrated by the tweet 
\textit{@user I hated Effexor. It makes you hungry, dizzy, and lethargic. That culminated in a large weight gain for me},
where \textit{hungry}, \textit{dizzy}, and \textit{lethargic} are a list of ADRs occurring as a phrase for the drug Effexor. As we found from preliminary experiments that with just attention it cannot detect all the ADR words in these phrases, we introduce \textit{coverage } into our decoder to solve this problem.   
For every decoder timestep we keep track of a coverage vector $c_t$, implemented as in \cite{see2017get} for summarization. The coverage vector sums up the attention distribution from all the previous decoder timesteps. However, unlike in \cite{see2017get} we impose a window for the previous decoder timesteps and set it to 3. We do this as most ADRs are comprised of few words and considering attention from all the previous timesteps can include attention from non-ADR words, which can jeopardize the attention distribution of ADR phrase located at the end of a post. With the coverage vector, in case if a word in a phrasal ADR does not get identified, the attentive words based on the attention distribution for the neighboring ADR words can help locate it. In other words, it helps secure the ADR label for the other words in a phrasal ADR. Coverage also contributes in avoiding mislabeling an Indication as an ADR and vice versa as it implicitly keeps track of the location of a word. That is, unlike conventional additive attention - which only uses encoder states $h_t$ to find the attentive words - by summing up attention values assigned so far, it provides information about how far the model has attended from the beginning of the sentence. This way it learns the boundaries of attention for the ADR and Indication words. It is defined as, 

\begin{equation}
c_t = \sum\limits_{s=t-1}^{s=t-3} a_s 
\end{equation}

With only attention mechanism, if for the word \textit{sleeping} in the first example attention was focused on the words \textit{Seroquel}, \textit{freaking} and \textit{out} to identify it as an ADR, it does not guarantee that it will attend to similar words for \textit{15} and recognize it as a part of the ADR phrase since \textit{15} can frequently appear with words irrelevant to ADR mention. By augmenting attention mechanism with \textit{coverage}, we hypothesize that by also focusing on the words that were attended to in the previous time steps, \textit{15} can be properly tagged as an ADR. That is, when decoding \textit{15} it will pay attention to the words \textit{Seroquel}, \textit{freaking} and \textit{out}, that were helpful in tagging \textit{sleeping} correctly.  Hence this hybrid attention with coverage module can enable capturing the words in an ADR phrase, which otherwise just with attention can get missed. 

We update the attention distribution to the following as coverage is now part of the attention mechanism,

\begin{equation}
{\textit{g}_t}_i = {\textit{v}_a^T}tanh\left(W_ah_i + W_bs_{t-1} + W_c{c_t}_i + b_{attn}\right)
\end{equation}

\subsection{Output}
Our task-specific output is calculated as the following,

\textbf{ADR Classification}:
The context vector can be viewed as a high-level representation of the post and is used as features for the classification task.
\begin{equation}
y = softmax\left(W_{class}{con} + b_{class}\right),
\end{equation}
where $W_{class}$ and $b_{class}$ are learnable parameters.

\textbf{ADR Labeling and Indication Labeling}:
The context vector at each decoder timestep, ${con}_i$, can be viewed as the final representation for the word, which is used with the decoder state, $s_i$, to predict the output at that timestep.

\begin{equation}
y_i = softmax\left(W'\left[s_i, {con}_i\right] + b'\right),
\end{equation}
where W' and b' are learned during training independently for the ADR labeling and Indication labeling tasks.
\section{Training the Model}
We train our model jointly over all the datasets. Cross-entropy is used as the loss function during training, which is defined for each task as:

\textbf{ADR Classification}:

\begin{equation}
L\left(\theta_{class}\right) = - \frac{1}{n}\sum\nolimits_x {\left( {{Y_x}\log {{Y'}_x} + \left( {1 - {Y_x}} \right)\log (1 - {{Y'}_x})} \right)},
\end{equation}
where n is the training data size, x means over all inputs, Y is the true label and Y' the predicted label probability. $\theta_{clas} = \left( \theta_{src}, \theta_1\right)$, where $\theta_{src}$ is a collection of shared parameters among all the tasks for the encoder and $\theta_1$ are the parameters for ADR Classification decoder. 

\textbf{ADR Labeling}:

\begin{equation}
L\left(\theta_{ADRlab}\right) = \frac{1}{n}\sum\nolimits_{x}{\log{P}\left(Y_x|X_x;\theta_{ADRlab}\right)},
\end{equation}
where $\theta_{ADRlab} = \left( \theta_{src}, \theta_2\right)$ and $\theta_2$ is the parameter for ADR Labeling decoder. 

\textbf{Indication Labeling}:

\begin{equation}
L\left(\theta_{INDlab}\right) = \frac{1}{n}\sum\nolimits_{x}{\log{P}\left(Y_x|X_x;\theta_{INDlab}\right)},
\end{equation}
where $\theta_{INDlab} = \left( \theta_{src}, \theta_3\right)$ and $\theta_3$ is the parameter for Indication Labeling decoder.

During training, we use the weighted loss which consists of a total number of T tasks:
\begin{equation}
total\hspace{0.1cm}loss = \sum\limits_{t=1}^{T} \omega_tL_t.
\end{equation}

Total loss is a linear combination of loss for all tasks, where $\omega_t$ is the weight for each task t respectively.

\section{Experimental Settings}
\subsection{Datasets}
\textbf{ADR Classification}: We use the Twitter dataset from PSB 2016 Social Media Shared Task for ADR Classification \cite{sarker2016social}. This dataset was created from tweets collected using generic and brand names of the drugs, along with their phonetic misspellings. It contains binary annotations of `0' and `1' referring to `ADR' and `notADR' respectively. Although the original dataset contains a total of 10,822 tweets, we could download only 7044 tweets from the given IDs. We consider only the 1081 tweets out of these that overlap with the tweets in ADR Labeling dataset as belonging to `ADR' class. Another 1081 tweets are randomly sampled from the `notADR' class. This is done as all the tasks share the same encoder and we exploit the shared input representations. We manually labeled the supplemental dataset discussed below and add it to the existing dataset. We divided the tweets randomly into training, test and validation datasets with splits of 70\%-15\%-15\%.

\textbf{ADR Labeling}: For ADR Labeling, we use the Twitter dataset from PSB 2016 Social Media Shared Task for ADR Extraction \cite{sarker2016social}. It contains around 2000 tweets annotated with tweet ID, start offset, end offset, semantic type (ADR/Indication), UMLS ID, annotated text span and the related drug. However, at the time of this study 1081 of the annotated tweets were available for download. We supplemented this dataset with a small dataset \cite{cocos2017deep} of 203 tweets collected between May 2015 to December 2015 from Twitter. We use the split 70\%-15\%-15\% for the training, test and validation datasets. We customized this dataset to only include the ADR annotations. 

\textbf{Indication Labeling}: The corpus and splits used for indication labeling are the same as that for ADR Labeling. We customized this dataset to only include the Indication annotations.

\subsection{Training Details}
Our implementation is based on the open source deep learning package Tensorflow \cite{abadi2016tensorflow}. Glove \cite{pennington2014glove} toolkit was used to pre-train the word embeddings, which were then used to initialize the embeddings in the model. We specifically used the model trained on Twitter in Glove to be able to generate the word-embeddings of the words unique to our Twitter dataset. We set the number of units in the LSTM cell to 128 and the dimensionality of word and char embeddings are set to 200 and 128 respectively. All the forget biases in the LSTMs are set to 1. In every epoch, we perform mini-batch training of each parallel task corpus with a batch size of 16. Regularization is done on the non-recurrent connections with a dropout rate of 0.5. We used Adam Optimization \cite{kingma2014adam}  method with a learning rate of 0.1 during training. All weights and biases in the attention and coverage component were initialized with Xavier \cite{glorot2010understanding} initialization. The development set was used to tune the value of the hyper-parameter weight, $\omega$, for the loss of each task. The setting 1, 1 and 0.1 for the losses of ADR labeling, Indication labeling, and ADR classification are used for the final experiments as it gave best results on the development set.

\section{Results and Discussion}
\subsection{Evaluation}
We use precision, recall and F-1 score as the evaluation measures. For a particular class $i$, precision and recall can be defined with the respective equations $p_i =  \frac{{TP}_i}{{TP}_i + {FN}_i}$ and $r_i = \frac{{TP}_i}{{TP}_i + {FP}_i}$ , where ${TP}_i$ is the number of true positives, ${FN}_i$ is the number of false negatives and ${FP}_i$ is the number of false positives. F-1 score is calculated from the precision and recall as $F\text{-}1 = \frac{2{p_i}{r_i}}{p_i + r_i}$.

For two tagging tasks, namely, ADR and Indication Labeling, we did approximate matching \cite{tsai2006various, cocos2017deep} on the predicted labels for the ADR phrasal words against their actual labels. Approximate matching works by checking if one or more of the ADR spans in an ADR phrase could be identified correctly with the `ADR' tags. For example, for the following tweet \textit{I was on Cymbalta for 5 days. Cold turkey had sweats, migraine, tremors while on \& 3 days after.} with the actual ADR span  \textit{sweats, migraine, tremors}, predicting the tag as `ADR' for any of the three spans or their combinations would be considered correct. The approximate match precision and recall are calculated as:\newline
\begin{equation}
p_i =  \frac{\# \text{ of ADR spans correctly tagged}}{\# \text{ of predicted ADR spans}}
\end{equation}
\begin{equation}
r_i =  \frac{\# \text{ of ADR spans correctly tagged}}{\# \text{ of actual ADR spans}}
\end{equation}

\subsection{Baselines}
As we could not find any previous work that performs multi-task learning on pharmacovigilance tasks, we compare it against two baseline methods and state-of-the-art approaches for the independent tasks to demonstrate the effectiveness of our proposed model. 
\begin{itemize}
\item \textbf{BLSTM-Random, BLSTM-Pretrained-learnable, BLSTM-Pretrained-fixed}: The architecture of this model is known as Bidirectional Long Short-Term Memory (BLSTM) RNN \cite{cocos2017deep}.  It combines a forward RNN and a backward RNN and uses only word embeddings as the features. In BLSTM-Random the word-embeddings are randomly initialized and treated as learnable parameters. BLSTM-Pretrained-learnable and BLSTM-Pretrained-fixed use pre-trained word-embeddings trained on a large non-domain specific Twitter dataset. The only difference between them is that BLSTM-Pretrained-learnable treats the words-embedding values as learnable parameters, while BLSTM-Pretrained-fixed as fixed constants. 
\item \textbf{CRNN}: State-of-the-art model for doing ADR Classification task. CRNN \cite{huynh2016adverse} is a convolutional neural network concatenated with a recurrent neural network. They used GRU as the basic RNN unit and RLU for the convolutional layer. 
\item \textbf{CNNA}: State-of-the-art model for doing ADR Classification task. CNNA \cite{huynh2016adverse} is a convolutional neural network with attention mechanism incorporated.   
\item \textbf{MT-NoAtten}: Our multi-task framework for pharmacovigilance tasks without any attention. We use a non-attention RNN as the decoder in this case.
\item \textbf{MT-Atten}: Our multi-task framework for pharmacovigilance tasks with just attention mechanism. Coverage is turned off during training.
\item \textbf{MT-Atten-Cov}: This is our proposed multi-task framework for pharmacovigilance tasks with combined attention and coverage mechanism.
\end{itemize}

\subsection{Overall Performance}
To show the validity of our model, we report results on three experiments. The results obtained from applying our model to the test sets are presented in Tables \ref{tab::test_results}, \ref{tab::adr_classification} and \ref{tab::adr_labeling}.

\begin{table*}[]
\centering
\caption{Test set results for the three tasks with the proposed model and the two baselines.}
\label{tab::test_results}
\begin{tabular}{l|lll|lll|lll}
\toprule
                    & \multicolumn{3}{l}{MT-Atten-Cov} & \multicolumn{3}{l}{MT-Atten Baseline} & \multicolumn{3}{l}{MT-NoAtten Baseline} \\
Metrics             & P (\%)        & R (\%)        & F-1 (\%)        & P (\%)           & R (\%)          & F-1 (\%)         & P (\%)           & R (\%)           & F-1 (\%)          \\ \midrule
ADR Classification  & ~N/A       & ~N/A      & ~N/A       & 72.88       & 70.54      & 70.69      & 69.63       & 60.60        & 55.62       \\
ADR Labeling        & 72.31     & 87.5     & 79.24     & 70.88       & 86.81      & 78.04      & 60.50       & 78.88       & 68.47       \\
Indication Labeling & 47.50     & 50.2     & 48.82     & 46.87       & 50.00      & 48.38      & 34.22       & 41.20       & 37.38      \\\bottomrule
\end{tabular}
\end{table*}

\begin{table}[]
\centering
\caption{Comparison of ADR Classification Task test results to previous approaches. Single-Atten-Cov Task refers to the independent task model trained on only the ADR Classification dataset.
}
\label{tab::adr_classification}
\begin{tabular}{llll}
\toprule
                     & P (\%) & R (\%) & F-1 (\%) \\ \midrule
CRNN                 & 49.00  & 55.00  &  51.00  \\
CNNA                 & 40.00  & 66.00  &  49.00     \\ 
Single-Atten-Cov     & 70.21  & 70.05  &  70.13  \\ 
MT-Atten             & \textbf{72.88}  & \textbf{70.54}  &  \textbf{70.69}  \\\bottomrule
\end{tabular}
\end{table}

\begin{table}[]
\centering
\caption{Comparison of ADR Labeling Task test results to previous approaches. Single-Atten-Cov Task refers to the independent task model trained on only the ADR Labeling dataset.}
\label{tab::adr_labeling}
\begin{tabular}{llll}
\toprule
                     & P (\%) & R (\%) & F-1 (\%) \\ \midrule
BLSTM-Random                & 64.57  & 63.32  &  62.72 \\
BLSTM-Pretrained-learnable                & 60.47  & 80.70  &  68.58  \\
BLSTM-Pretrained-fixed                & 70.43  & 82.86  &  75.49  \\ 
Single-Atten-Cov  & 71.50  & 86.22  &  78.17  \\
MT-Atten-Cov             & \textbf{72.31}  & \textbf{87.50}  &  \textbf{79.24}  \\\bottomrule
\end{tabular}
\end{table}

In the first experiment, we train our multi-task model jointly on the three parallel datasets corresponding to each of the three tasks. Results for each task from this experiment are compared against two baselines, which are depicted in Table 1. As the purpose of coverage is to be able to have a greater coverage for the ADR/ Indication words in an ADR/ Indication phrase, we turn it off in the classification decoder and use it in the two tagging tasks. So, results for classification task are not reported with MT-Atten-Cov model. For the remaining two tasks, we can observe that our method outperforms both the baselines in terms of the precision, recall, and F-score. Although approximate matching would consider the identification of any ADR/ Indication span as a true positive and we would expect comparable results for MT-Atten-Cov and MT-Atten models, the fact that MT-Atten-Cov has superior results empirically confirms that incorporating coverage with attention helps in capturing `phrasal' and single ADR words which are not attended to with just the attention mechanism. We achieve a 1.50\% and 0.90\% F-1 score improvement for ADR labeling and Indication labeling respectively over the MT-Atten model. The experimental values for Indication labeling task happen to be low across all the models due to the sparsity of tweets containing indication words. Nevertheless, we gain improvement with our model against both the baselines. We gain an F-1 score improvement of 21. 32\%, 13.59\% and 23.43\% over the MT-NoAtten model for classification, ADR detection, and Indication labeling respectively. By examining the precision and recall results, we can say that the better performance of our model can be attributed to the improvement of both. 

In the second and third experiments, we train our model separately on ADR classification and ADR detection tasks respectively and provide the comparison against results from the model trained in experiment one. As Indication detection was not performed as an independent task in any previous works, we do not provide a separate table with comparisons to previous approaches for it. We can see from Table 2 that our multi-task model improves the performance in terms of the F-1 score with 0.79 \% for classification compared to a single task model. While for the ADR detection task results in Table 3, it makes 1.35\% improvement. These empirical findings cast light on that shared input representations and interactions between tasks result in mutual benefit for all tasks. Comparing all the independent classification models (Single-Atten-Cov Task, CRNN, and CNNA) against each other, we can further see the advantage of using attention for the classification task, where Single-Atten-Cov has an average improvement of 28.71\% over CRNN and CNNA. Although CNNA incorporates attention in its model, we assume using an RNN encoder-decoder with attention is more helpful. Similarly for ADR detection, from Table 3 we can see that our single task ADR detection model (Single-Atten-Cov Task) outperforms the best model among BLSTM (BLSTM-Pretrained-fixed) by 3.43\%. Classification performance for our multi-task model attained higher F-1 scores over the CRNN and CNNA models by 27.85\% and 30.66\% respectively, while for ADR detection, multi-task model improves by 4.73\% over the best performing BLSTM models. 

\subsection{Case Study}
To get a deeper insight into how augmenting coverage to attention mechanism benefits our model, we sample several tweets from the test dataset. The tags predicted by our model for these tweets are compared against those from the baseline MT-Atten model. All usernames are anonymized for privacy concerns. In order to validate the results that our model produced, we further visualize the attention heatmaps for some of the tweets depicted in Figure 3. 

The following two tweets illustrate the ability of our model to correctly label the single ADR word as `ADR', while the MT-Atten model makes wrong prediction. This justifies the higher precision and recall gained by our model over MT-Atten.
\begin{itemize}
\item \textbf{Tweet 1}: \textit{@user1 bloody zombie I also take Venlafaxine thats for cronic depression bipolar is a difficult illness to deal with}\\
\textbf{True Tags}: \textit{[`O', `O', `ADR', `O', `O', `O', `O', `O', `O', `O', `O', `O', `O', `O', `O', `O', `O', `O', `O']}\\
\textbf{MT-Atten}: \textit{[`O', `O', `O', `O', `O', `O', `O', `O', `O', `O', `ADR', `O', `O', `O', `ADR', `O', `O', `O', `O']}\\
\textbf{MT-Atten-Cov}: \textit{[`O', `O', `ADR', `O', `O', `O', `O', `O', `O', `O', ' O ', `O', `O', `O', `O', `O', `O', `O', `O']}\\
\textbf{ADR}: zombie
\end{itemize}

The degree of \textit{attention} received by each source word when decoding a target word for Tweet 1 is visualized in Figure 3, where darker shades denote higher score. When it is predicting the tag for target token `zombie', more attention is given to the words `zombie', `Venlafaxine', `difficult' and `illness'. Due to coverage, attentive words `difficult' and `illness' from previous timesteps were also attended. These words facilitated in predicting the `ADR' tag for this word. Furthermore, \textit{indication} words `bipolar' and `depression' receiving low attention weights are prevented from mislabeling.
\begin{itemize}
\item \textbf{Tweet 2}: \textit{Cymbalta, my mood has worsened}\\
\textbf{True Tags}: \textit{[`O', `O', `ADR', `O', `O']}\\
\textbf{MT-Atten}: \textit{[`O', `O', `O', `O', `ADR']}\\
\textbf{MT-Atten-Cov}: \textit{[`O', `O', `ADR', `O', `ADR']}\\
\textbf{ADR}: mood
\end{itemize}
\begin{figure*}[h!]
\includegraphics[width=7in]{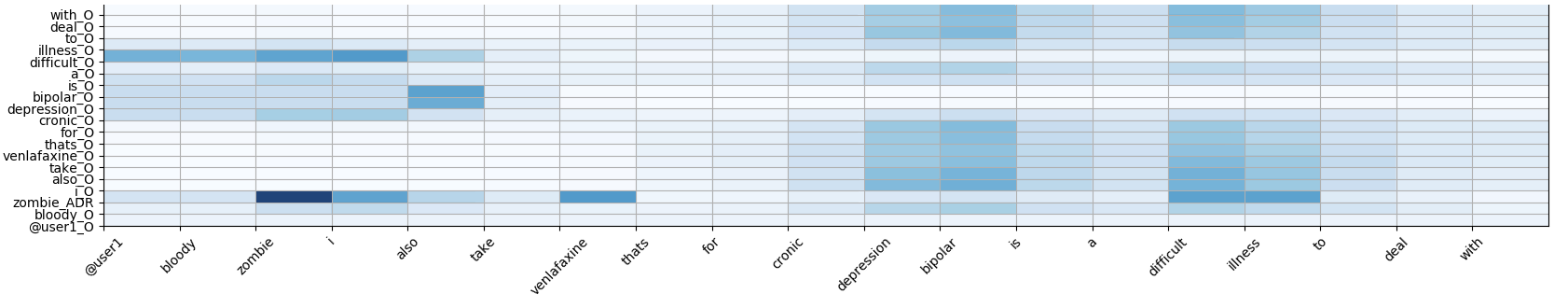}
\includegraphics[width=7in]{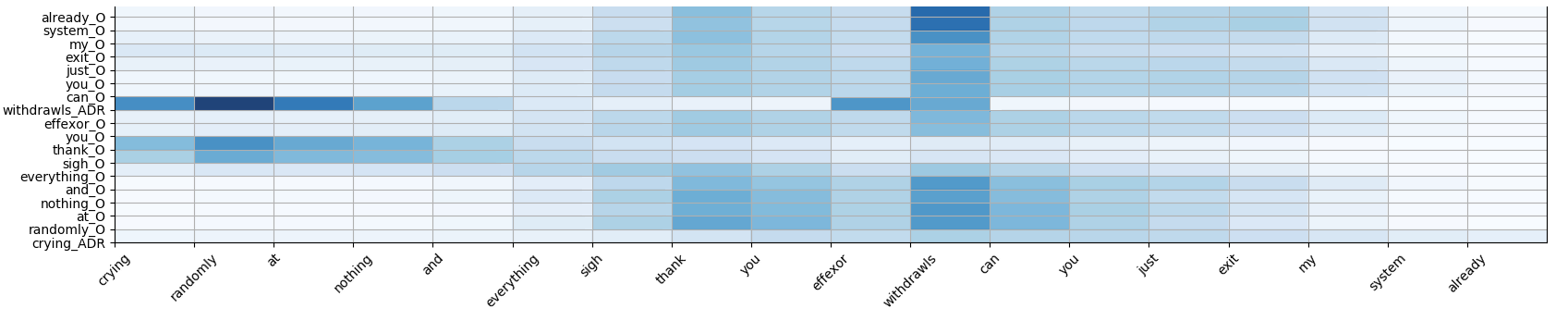}
\includegraphics[width=7in]{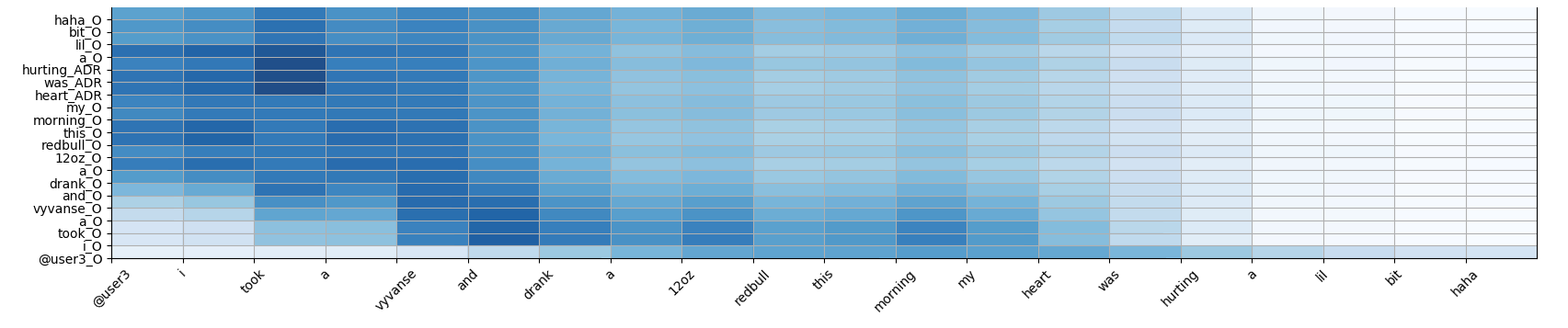}
\caption{Attention Heatmaps for Tweets 1, 4 and 5.}\label{fig::atttention_twitter}
\end{figure*}

The following tweets contain ADRs captured by our model but missed by MT-Atten.
\begin{itemize}
\item \textbf{Tweet 3}: \textit{I'm sorry you have ostrich halitosis @user2 but that IS a side effect of once monthly boniva.}\\
\textbf{True Tags}: \textit{[`O', `O', `O', `O', `O', `O', `ADR', `O', `O', `O', `O', `O', `O', `O', `O', `O', `O', `O']}\\
\textbf{MT-Atten}: \textit{[`O', `O', `O', `O', `O', `O', `O', `O', `O', `O', `O', `O', `O', `O', `O', `O', `O', `O']}\\
\textbf{MT-Atten-Cov}: \textit{[`O', `O', `O', `O', `O', `O', `ADR', `O', `O', `O', `O', `O', `O', `O', `O', `O', `O', `O']}\\
\textbf{ADR}: halitosis
\end{itemize}
\begin{itemize}
\item \textbf{Tweet 4}: \textit{Crying randomly at nothing and everything. Sigh. Thank you \#effexor \#withdrawls Can you just exit my system already.}\\
\textbf{True Tags}: \textit{[ADR', `O', `O', `O', `O', `O', `O', `O', `O', `O', `ADR', `O', `O', `O', `O', `O', `O', `O']}\\
\textbf{MT-Atten}: \textit{[`O', `O', `O', `O', `O', `O', `O', `O', `O', `O', `O', `O', `O', `O', `O', `O', `O', `O']}\\
\textbf{MT-Atten-Cov}: \textit{[`O', `O', `O', `O', `O', `O', `O', `O', `O', `O', `ADR', `O', `O', `O', `O', `O', `O', `O']}\\
\textbf{ADR}: Crying
\textbf{ADR}: withdrawls
\end{itemize}

From the attention heatmap of this tweet shown in Figure \ref{fig::atttention_twitter}, we can see that the first ADR word 'crying' has overall low attention for all the words. Also, as it happens to be the first word of the sentence, no coverage was passed. This is the reason why it is not detected as an ADR by both the baseline and our model. Whereas, for the second ADR `withdrawals' source words `crying', `randomly', `at' and `nothing' were attended to more than others. Moreover, `Effexor' and `withdrawals' from the attention of previous word `Effexor' were also instrumental in the right tag assignment of `withdrawals'. 

With the following tweet we can see how \textit{coverage} causes attention from previous ADR word 'heart' to identify 'was' as part of the ADR phrase. We can see the baseline model MT-Atten mislabels it.
\begin{itemize}
\item \textbf{Tweet 5}: \textit{ @user3 I took a vyvanse and drank a 12oz redbull this morning. My heart was hurting a lil bit haha}\\
\textbf{True Tags}: \textit{[`O', `O', `O', `O', `O', `O', `O', `O', `O', `O', `O', `O', `O', `O', `ADR', `ADR', `ADR', `O', `O', `O', `O']}\\
\textbf{MT-Atten}: \textit{[`O', `O', `O', `O', `O', `O', `O', `O', `O', `O', `O', `O', `O', `O', `ADR', `O', `ADR', `O', `O', `O', `O']}\\
\textbf{MT-Atten-Cov}: \textit{[`O', `O', `O', `O', `O', `O', `O', `O', `O', `O', `O', `O', `O', `O', `ADR', `ADR', `ADR', `O', `O', `O', `O']}\\
\textbf{ADR}: heart was hurting
\end{itemize}
\section{Related Work}
\textbf{Pharmacovigilance in Social Media}: Pharmacovigilance has become a very active area of research \cite{world2002importance}. Initial work on pharmacovigilance in social media performed detection and extraction tasks on health forum (e.g., DailyStrength, MedHelp) data \cite{patki2014mining,leaman2010towards, nikfarjam2011pattern}. These works also focused on investigating posts mentioning adverse reactions associated with a fewer number of drugs. Over the years, a wide range of drugs have been used and other platforms such as Twitter have emerged as a valuable source for ADR monitoring \cite{freifeld2014digital, o2014pharmacovigilance, ginn2014mining}. 

Two of the most prevalent tasks for pharmacovigilance studies are the detection of posts containing ADR mention and extraction of the ADR mentions from posts \cite{sarker2015utilizing}. Due to the scarcity of annotated resources for this study, most works have been performed using small annotated datasets \cite{leaman2010towards, nikfarjam2011pattern}. While some work used large unannotated dataset employing an unsupervised approach \cite{chee2011predicting, benton2011identifying}. For ADR extraction, ADR lexicons and knowledge bases have been the most widely used resource \cite{sarker2015utilizing}. However, medical terms are rarely used in social media posts and raise mapping issues. Applying deep neural networks for pharmacovigilance has found its way in some recent works \cite{cocos2017deep, huynh2016adverse}.  These approaches benefit from not having to explicitly specify the features, but rather learn them in the training process. 

As mention of `Indication' co-occurs with ADRs in a post, no previous work has performed `Indication' detection/extraction as a separate task. Although some works have used it as features in a machine learning approach \cite{nikfarjam2015pharmacovigilance}, the sparsity of `indications' has been an issue. With our multi-task approach that includes `Indication' extraction as a separate task, we show that interaction between related tasks can improve learning for all.

\noindent\textbf{Multi-task Learning}: Use of multi-task learning models has become ubiquitous for many machine learning applications in areas ranging from natural language processing, speech recognition to computer vision \cite{ruder2017overview}. Multi-task learning with encoder-decoder come in one of the three flavors: one-to-many, many-to-one and many-to-many approaches \cite{luong2015multi}. In One-to-many models, tasks share a common encoder and have a task-specific decoder. A one-to-many approach was used in \cite{dong2015multi} for translation task from a source language to multiple target languages. The shared encoder captured the syntactic and semantic similarity existing across different languages. On the other hand, a many-to-one approach is suitable for tasks where a decoder can be easily shared, such as multi-source translation. Lastly, a many-to-many approach allows multiple encoders and decoders. 

Our one-to-many approach is similar in spirit to \cite{liu2016attention} used for joint slot filling and intent detection tasks. However, our work has the following differences: 1) learns more tasks jointly 2) use of coverage with attention mechanism. 3) use of pre-trained word embeddings.

\section{Conclusion}
Performing pharmacovigilance on Twitter can augment the existing ADR surveillance system which suffers from various clinical limitations. In this work, we provide an end-to-end solution for three ADR detection and extraction tasks. These problems have conventionally been approached separately and didn't leverage the interactions between the tasks. Exploiting the similarity that these tasks have, we proposed a multi-task encoder-decoder framework. All the tasks share one encoder to model the interactions and semantic/syntactic similarity between them. While each has its own decoder to produce output specific to that task. Our empirical findings validate how learning all tasks jointly improves precision and recall over state-of-the-art approaches. Additionally, the proposed solution is a hybrid attention model with coverage to deal with ADRs occurring as phrases. Through results and case studies we show that not only is this hybrid model able to achieve higher phrasal ADR word coverage compared to the baselines but is also able to identify single ADRs correctly. 
\section{Acknowledgments}
This work is supported in part by NSF through grants IIS-1526499, CNS-1626432 and NSFC 61672313.

\balance

\end{document}